\documentclass[journal,twoside]{IEEEtran}

\usepackage[pdftex, pagebackref, colorlinks]{hyperref}
\usepackage{bookmark}

\usepackage{graphicx}
\usepackage{overpic}
\usepackage{tabularx}

\usepackage{amsmath,amssymb} % define this before the line numbering.
\usepackage{algorithm,algpseudocode}
\usepackage{multirow}

\begin{document}
\title{Inner and Inter Label Propagation: \\
Salient Object Detection in the Wild
}

\author{Hongyang~Li,~\IEEEmembership{Student Member,~IEEE,}
        Huchuan~Lu,~\IEEEmembership{Senior Member,~IEEE,} \\
        Zhe~Lin,~\IEEEmembership{Member,~IEEE,}
        Xiaohui~Shen,~\IEEEmembership{Member,~IEEE,}
        and~Brian~Price,~\IEEEmembership{Member,~IEEE}

\thanks{
%Manuscript received April 19, 2005; revised December 27, 2012.
This work was supported by the
Natural Science Foundation of China \#61472060 and the Fundamental Research Funds for Central Universities under
Grant DUT14YQ101. We thank Lijun Wang for helpful discussions along the way; Na Tong and Wei Liu for careful proofreading and reviewers for fruitful and detailed comments.}
\thanks{
H. Li is affiliated with the Department
of Electronic Engineering, The Chinese University of Hong Kong.
This research project was done when he was a final-year undergraduate and
served as a research assistant in the computer vision group at
Dalian University of Technology.
E-mail: hongyangli2020@gmail.com.}
\thanks{H. Lu is with the School of Information and Communication
Engineering, Dalian University of Technology. Email: lhchuan@dlut.edu.cn.}
\thanks{
Z. Lin, X. Shen and B. Price are with Adobe Research, San Jose, USA. Emails: \{zlin, xshen, bprice\}@adobe.com
}
}

\markboth{IEEE transactions on image processing, 2015%, Vol. xx, No. xx, xx 2015
}
{H.Y. Li, \MakeLowercase{\textit{et al.}}:    Inner and Inter Label Propagation: Salient Object Detection in the Wild}


\maketitle

\begin{abstract}
~In this paper, we propose a novel label propagation based method for saliency detection.
%A key observation is that the object and background can be estimated via labels extracted from them.
A key observation is that saliency in an image can be estimated by propagating the labels extracted from the most certain background and object regions.
For most natural images, some boundary superpixels serve as the background labels  and the saliency of other superpixels are determined by ranking their similarities to the boundary labels based on an inner propagation scheme.
For images of complex scenes, we further deploy a 3-cue-center-biased
objectness measure to pick out and propagate foreground labels.
A co-transduction algorithm is devised to fuse both
boundary and objectness labels based on an inter propagation scheme.
The compactness criterion decides whether the incorporation of objectness labels is necessary,
thus greatly enhancing computational efficiency.
Results on five benchmark datasets with pixel-wise accurate annotations show that the proposed method achieves superior performance compared with the newest
state-of-the-arts in terms of different evaluation metrics.
\end{abstract}


% \fi
\IEEEpeerreviewmaketitle

\section{Introduction}\label{sec:intro}

%\IEEEPARstart{T}{his} demo file is intended to serve as a ``starter file''
%for IEEE journal papers produced under \LaTeX\ using
%IEEEtran.cls version 1.8 and later.
%% You must have at least 2 lines in the paragraph with the drop letter
%% (should never be an issue)
%I wish you the best of success.

\IEEEPARstart{H}{umans} have the capability to quickly prioritize external visual stimuli and localize their most interested regions in a scene \cite{UFO13}.
In recent years, visual attention has become an important research problem in both
neuroscience and computer vision.
One branch focuses on eye fixation prediction \cite{IT98,Judd_2009,Boolean} to
investigate the mechanism of human visual systems whereas the other trend
concentrates on salient object detection \cite{11pami/Liu_Learning,08cvs/achanta_salient,gopalakrishnan2009salient} to accurately identify a region of interest.
Saliency detection has served as a pre-processing procedure for many vision tasks, such as collages \cite{10cvpr/goferman_context}, image compression \cite{Compress}, stylized rendering \cite{RC11}, object recognition \cite{recog}, visual tracking \cite{tracking}, image retargeting\cite{Hyper13}, etc.

In this work, we focus on the salient object detection.
%Due to its importance, vast efforts have been made to explore theoretically plausible and computationally effective algorithms.
%
Recently, many low-level features directly extracted from images %,\emph{ e.g.}, color, luminance, gradient,
have been explored.
It has been verified that color contrast is a primary cue for satisfying results \cite{11pami/Liu_Learning,RC11}.
% luminance \cite{06LC},
Other representations based on the low-level features try to exploit the intrinsic textural difference between the foreground and background, including focusness \cite{UFO13}, textual distinctiveness \cite{TD13}, and structure descriptor \cite{PISA13}.
They perform well in many cases, but can still struggle in complex images.
Instead, we observe that the primitive appearance information alone is good enough to reflect the textural difference from the boundaries of superpixels.

Due to the shortcomings of low-level features, many algorithms have turned to incorporating higher-level features \cite{Northwestern12,DRFI13,Submodular13,US13}.  One type of higher-level representations that can be employed is the notion of objectness \cite{Objectness}, or how likely a given region is  an object.
For example, Jiang \emph{et al.} \cite{UFO13} compute a saliency measure by combining the objectness values of many overlapping windows.
However, %a take-all-cue mechanism from the original objectness computation will suffer from inaccuracy due to different testing datasets.
%Besides,
using the objectness measure directly to compute saliency may produce unsatisfying results in complex scenes when the objectness score fails to predict true salient object regions \cite{SVO11,OB13}.
%Therefore, we consider  objectness computed from only three cues  as hint labels of the foreground.
A better way to employ  high-level objectness is to consider the scores as hints of the foreground.

To this end, we put forward a unified approach to incorporate low-level features and the objectness measure for saliency detection via label propagation.
Since the border regions of the image are good indicators to distinguish salient objects from the background \cite{Manifold,Markov_dlut},
we observe that the boundary cues can be used to estimate the appearance of the background while the objectness cues focus on the characteristics of the salient object. Therefore, a refined co-transduction \cite{CoTrans} based method, namely \emph{label propagation saliency} (LPS), is proposed.  %to incorporate both boundary and objectness cues.
In this framework, the most certain boundary and object regions are able to propagate saliency information in order to best leverage their complementary influence.  As the boundary cue can be quite effective in some cases and the objectness measure requires additional computation, a compactness criterion is  further devised to determine whether the  results propagated by boundary labels are sufficient.%already sufficiently good.

%\vspace{-.5cm}
\begin{figure*}
  \centering
  \includegraphics[width=\textwidth]{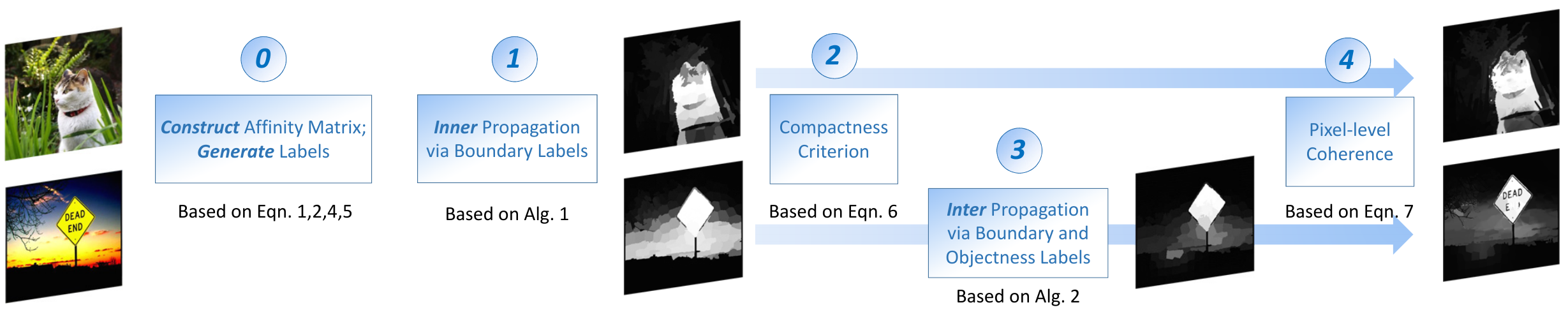}
  \vspace{-.2cm}
%\footnotesize{
  \caption{% \small
  Pipeline of the label propagation saliency algorithm.
  First, we construct the normalised affinity matrix from the superpixels and generate boundary and objectness label sets, respectively; then the inner propagation is conducted to have initial saliency maps; third, the compactness criterion chooses those who need a further refinement by the inter propagation scheme; finally, all maps are enhanced via a pixel-level saliency coherence.
  }\label{pipeline}
  %}
\end{figure*}
%\vspace{-.5cm}
%Combining boundary  with objectness labels on both regional and pixel-wise level,

Fig.\ref{pipeline} shows the pipeline of our method.
First, we extract the affinity matrix and choose some border nodes as  labels to represent the background (Sec.\ref{sec:affinity}).
The \emph{inner} propagation is implemented to obtain the regional maps (Sec.\ref{sec:inner}). Second,
a compactness criterion  is introduced to evaluate whether
these maps need a further refinement %using the objectness measure
(Sec.\ref{sec:co-trans}). Third, the  \emph{inter} propagation incorporates objectness labels
via a co-transduction algorithm to regenerate maps for images that fail  to work in the inner stage (Sec.\ref{sec:object}, \ref{sec:co-trans}). Fourth, all maps are updated at a
pixel level to  achieve coherency of the  saliency assignment (Sec.\ref{sec:coherence}).
%get better saliency coherence.
%
The contributions of our work include:
\vspace{-.1cm}
\begin{enumerate}

  \item A simple and efficient label propagation algorithm via boundary labels for most natural images  based on the reconstructed affinity matrix;
  \item A novel co-transduction framework to incorporate foreground labels obtained from the objectness measure with boundary cues for complex images;
  \item A compactness selection mechanism to decide whether the initial maps need an update, thus facilitating the computational efficiency.
\end{enumerate}
\vspace{-.1cm}

The experimental results show that the proposed method achieves superior performance in various evaluation metrics against other 27  state-of-the-arts
on five  image benchmarks.
Finally, the results and code
%will be shared upon acceptance.
are  shared for research
purposes\footnote{\href{https://github.com/hli2020/lps\_tip15}{
\texttt{ https://github.com/hli2020/lps\_tip15}}}.

%\hfill December 27, 2012

\section{Related Work}

Saliency estimation methods can be explored from different perspectives.
Basically, most works employ
a bottom-up approach via low level features while a few
incorporate a top-down solution driven by specific tasks.
%These methods can be roughly
%classified into biologically inspired solutions,
%computationally oriented approaches, or a combination \cite{09cvpr/Achanta_FTSaliency}.
%
%In general, all methods employ a low-level or bottom-up approach by determining contrast of image regions relative to their surroundings. Global contrast captures the holistic rarity or uniqueness from an image. However, when a foreground region is globally compared with the remaining of the scene, which inevitably includes other foreground regions, its contrast with the background is less distinct and the salient object is unlikely to be uniformly highlighted. Local contrast, on the other hand, accords with the neuroscience principle that neurons in the retina are sensitive to regions which locally stand out from their surroundings.
%
Early researches address saliency detection via biologically inspired models, such as Gaussian pyramids \cite{IT98},
fuzzy growing \cite{03ACMMM/Ma_Contrast-based}, graph-based activation \cite{07ANIPS/harel_graph}.
Other studies employ frequency domain methods \cite{07cvpr/hou_SpectralResidual,09cvpr/Achanta_FTSaliency,Freq11} to
determine saliency according to the spectrum
of the image's Fourier transform.
However, the results of these methods
exhibit undesirable blurriness and tend to highlight object boundaries rather than its entire area.

Recently, the saliency detection community has witnessed a blossom of high accuracy results under
distinctive frameworks \cite{AliECCV12}. Learning methods \cite{US13,DRFI13,Hyper13} integrate both low and high level
features to compute saliency based on parameters
trained from sample images. Although learning mechanisms perform well in proposing bounding boxes, they suffer in salient object detection due to the complex scenes of the background.
Shen \emph{et.al} \cite{Northwestern12} introduce high-level priors to form high-dimensional representations of the image and construct saliency in a low rank framework. Despite the complicated configuration, the resultant maps have unsatisfying saliency assignment near the salient object.

Faced with  the above issues and considering the limited knowledge of structural description mentioned in Sec.\ref{sec:intro}, we try to extract features in a simple and effective way.
Jiang \emph{et al.} \cite{Markov_dlut} introduce an absorbing Markov chain method where
the appearance divergence and spatial distribution between salient objects and the background are considered.
%
%A boolean map approach was proposed in \cite{Boolean}, which analyzes the topological structure of randomly generated boolean maps.
%Li \emph{et al.} \cite{PCA_dlut} defined  saliency measure as the reconstruction errors of PCA and sparse representation and fused two methods in a Bayesian framework.
Cheng \emph{et al.} \cite{RC11} formulate a regional contrast based
saliency  algorithm which simultaneously evaluates global and local contrast differences. %and spatial coherence.
Inspired by these works, we construct an affinity matrix based on the color feature of superpixels with two adjustments to involve spatial relations.

A novel label propagation method is proposed in \cite{GraphTrans}  to rank the similarity of data points to the query labels
for shape retrieval. We apply and refine the theory
to make full use of the background and foreground superpixels, which has been rarely studied in saliency detection.
Distinct from the work of Yang \emph{et al.} \cite{Manifold} where a manifold ranking algorithm assigns saliency
 based on priors of all boundary nodes,
in this work, (a) we only take some boundary nodes to eliminate salient regions that appear at the image border;
(b) both boundary and foreground nodes are selected as complementary labels in a co-transduction framework to fully distinguish salient areas from the background;
and (c) the revised label propagation algorithm has zero parameter
whereas in \cite{Manifold} the sensitive $\alpha$ has a vital effect on results in different datasets.

\section{The Label Propagation Algorithm}
We first introduce the  construction of the affinity matrix  in Sec.\ref{sec:affinity}, which is of vital importance during the label propagation. Then the inner propagation via boundary labels is proposed in Sec.\ref{sec:inner}. An objectness measure is
utilised to locate foreground labels in Sec.\ref{sec:object}. Sec.\ref{sec:co-trans} illustrates the co-transduction algorithm which takes into consideration
both boundary and objectness cues and the compactness criterion to classify initial maps generated from the inner propagation.
Finally, we refine the regional maps on pixel level to achieve saliency coherency in Sec.\ref{sec:coherence}.
%We employ a label propagation mechanism \cite{GraphTrans} to measure the similarity
%of superpixels to the query labels in a graph structure.
%The boundary and the objectness superpixels \cite{Objectness} are utilized to serve as the
%query labels respectively. A compactness score is proposed to
%evaluate whether the saliency map produced by boundary propagation
%needs a further refinement. The final saliency
%map is simply a production of the boundary propagation or a fusion outcome of the boundary and objectness
%propagation via a co-transduction algorithm \cite{CoTrans}.

\subsection{Affinity Matrix Construction}\label{sec:affinity}
We first construct an affinity matrix among superpixels to be used in the propagation algorithm. $L_0$ gradient minimization \cite{smooth} is implemented to obtain a soft abstraction layer while keeping vital details of the image.
Superpixels are generated to segment the smoothed image into $N$ regions by the SLIC algorithm \cite{slic},
where regions at the image border form a set of boundary nodes, denoted as $B$.
In this work, we refer the superpixel as a node or a region.

\begin{figure}
  \centering
  \includegraphics[width=6.5cm]{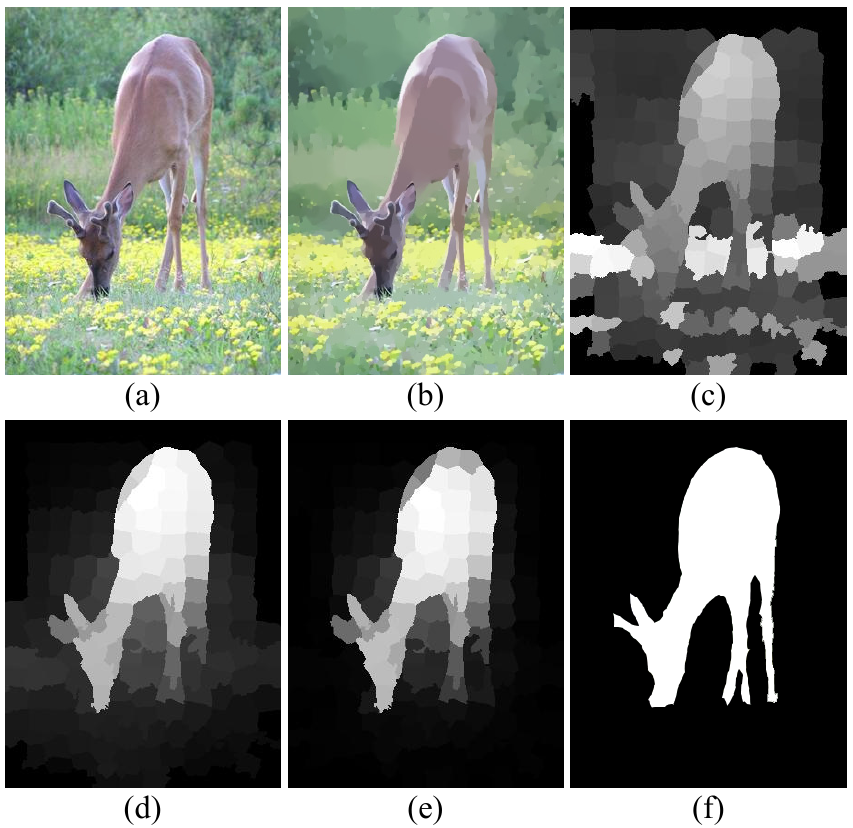}
  %\vspace{-.3cm}
    \caption{ Effects of affinity construction. (a) Input image;
  (b) $L_0$ smoothing; (c) Full connection; (d) No geodesic constraint;
  (e) Geodesic constraint; (f) Ground truth. }\label{geodesic}
  \end{figure}

The similarity of two nodes is measured by a defined %$L_2$-norm
distance of the mean features in each region.
Based on the intuition that neighboring regions are likely to share similar appearances and that remote ones
do not bother to have similar saliency values even if the appearance of them
are highly identical, we define the affinity entry $w_{ij}$ of superpixel $i$ to a certain node $j$ as:
%\vspace{-.5cm}

\begin{equation}\label{affinity}
  w_{ij} = \begin{cases}\exp(- \frac{D(\textbf{f}_i,   \textbf{f}_j)}{\sigma^{2}})  &   \textrm{  $j \in \mathcal{N}(i)$ or $i,j \in B$   }  \\
  0  & \textrm{ $i=j$ or otherwise }
  \end{cases}
  %\vspace{-.5cm}
\end{equation}
where $\textbf{f}_i, \textbf{f}_j$ denote the mean feature vectors of pixels inside node $i,j$ respectively,
$\sigma$ is
a tuning parameter to control strength of the similarity, $\mathcal{N}(i)$
%includes nodes that directly borders on node $i$ or share common boundaries of some regions with node $i$.
indicates the set of the direct neighboring %nodes that touch the boundary
nodes of superpixel $i$, as well as the direct neighbors of those neighboring nodes.
Therefore, we have an affinity matrix $\textbf{W}=[w_{ij}]_{N\times N}$
to indicate the similarity between any pair of superpixels,
a degree matrix
$\textbf{D}=diag\{ d_1,\dots, d_N \}$ where $d_i=\sum_j w_{ij}$ to
sum the total entries of each node to other nodes, and a row-normalized
affinity matrix:
\begin{equation}\label{affinity_eqn}
  \textbf{A} = \textbf{D}^{-1}\cdot \textbf{W}
\end{equation}
to be finally adopted.

Different from the common practice of a fully connected network among superpixels \cite{RC11,OB13}, there are two adaptions to construct the affinity entry in Eqn.\ref{affinity}.
First, a conception of $k$-layer neighborhood (here $k=2$) in graph theory is introduced.
The enlarged neighbors of the region enforce a spatial relationship that salient object tends to be clustered
rather than to be scattered.
%\vspace{-.3cm}
%
%\vspace{-.4cm}
Second, we adopt a geodesic constraint mechanism \cite{Markov_dlut,Manifold} to further enhance
the relationship among boundary nodes, \emph{i.e}., any two superpixels in $B$ are connected.
Since the boundary nodes serve as propagation labels, a strong connection among them
could better distinguish the background from the salient object.
%Besides, such a mechanism
%works well when salient object appears at the image boundary (see Fig.\ref{boundary}(c)).

The effects of affinity construction are illustrated in Fig.\ref{geodesic}.
We note that under a fully connected scheme in Fig.\ref{geodesic}(c), yellow flowers in the background are salient due to the mere
consideration of color and ignorance of spatial distance. Without a geodesic constraint scheme,
the saliency map in Fig.\ref{geodesic}(d) has vast background areas with low saliency assignment which leads to a low precision at a high recall in the precision-recall curve.

%\IEEEpubidadjcol

\begin{figure}
  \centering
  \includegraphics[width=8.7cm]{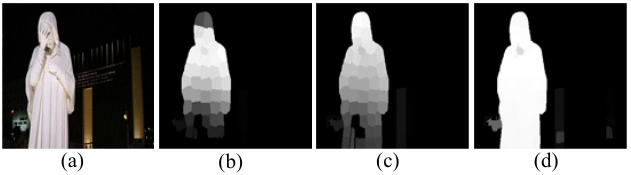}
  \caption{ Salient objects at image border.
  (a) Input image; (b) Use all boundary nodes; (c) No geodesic constraint;
  (d) Selected boundary nodes.
  %(e) Ground truth.
  }\label{boundary}
 \vspace{-.5cm}
\end{figure}

\subsection{Inner Propagation via Boundary Labels}\label{sec:inner}
Given an affinity matrix, we endeavor to propagate the information of the background labels  to estimate saliency measure
of other superpixels.
A shape similarity method that exploits the
intrinsic relation between labelled and unlabelled objects
is proposed in \cite{GraphTrans} to tackle the image retrieval problem via label propagation.
Given a dataset $R=\{ r_1, \dots, r_l, r_{l+1}, \dots, r_N  \}\in\mathbb{R}^{D\times N}$,
where the former $l$ regions serve as query labels and $D$ denotes the feature dimension, we seek out
a function $\textbf{V}=[V(r_1),\dots,V(r_N)]^T$ such that $\textbf{V}:R\rightarrow [0,1]\in\mathbb{R}^{N\times 1}$
indicates the possibility of how similar each data point is to the labels.
The similarity measure $V(r_i)$ satisfies
%\vspace{-.3cm}
\begin{equation}\label{propagate}
  V_{t+1}(r_i)=\sum_{j=1}^{N}a_{ij}V_{t}(r_j)
\end{equation}
where $a_{ij}$ is the affinity entry defined in Eqn.\ref{affinity_eqn} and
$t$ is the recursion step.
%The convergence verification of Enq.\ref{propagate} can be found in \cite{GraphTrans}.

\begin{figure}
  \centering
  \includegraphics[width=8cm]{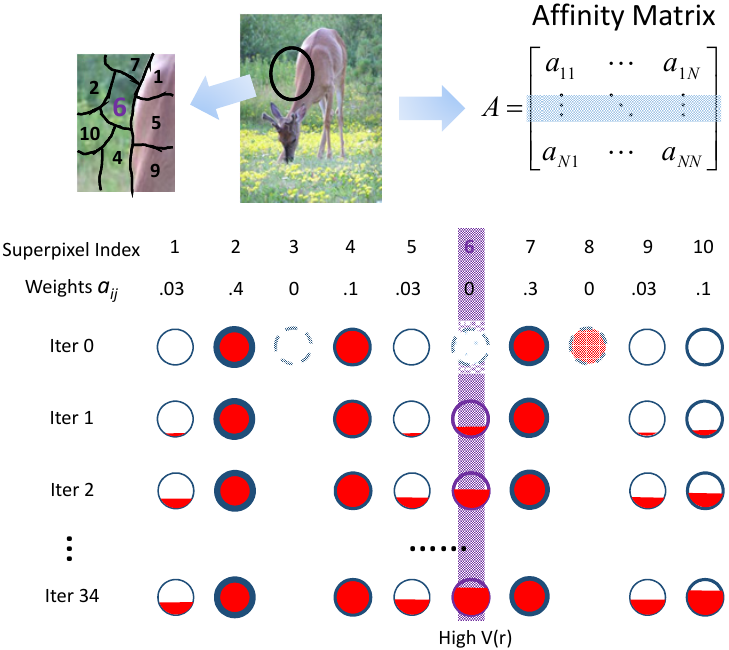}
  %\vspace{-.3cm}
  \caption{
  A toy example to illustrate how the inner propagation algorithm (Alg.\ref{alg1}), or Eqn.\ref{propagate}, works.
  For simplicity, we investigate one superpixel region ($\#6$) and see how its value $V(r)$ changes during each iteration.
  Assuming we have 10 regions in total and the 6-\textit{th} row of the normalised affinity matrix (weights $a_{ij}$) shows the similarity between the considered region and other regions. The dash-outline regions ($\#3,6,8$) are not neighbours
  of region 6 and thus not considered in the propagation. The outline weight of each circle indicates the affinity weight, \textit{i.e.}, the thicker it looks, the bigger $a_{ij}$ is.
  The red area inside each circle denotes the value of $V(r)$, since we assume region 2,4,7 are background labels,
  they have red colour fully filled within their circles in each iteration.
   }\label{toy}
\end{figure}

The similarity measure of query labels is fixed to be 1 during the recursive process
and the initial measure of unlabelled objects is set to be 0.
For a given region, the similarity $V(r_i)$  is learned iteratively via  propagation
of the similarity measures of its neighbors $V(r_j)$ such that
a region's final similarity to the labels is
effectively influenced by the features of its surroundings.
In other words, the new similarity will be large
iff all points $r_j$ that resemble $r_i$ are also quite similar to query labels.
Fig.\ref{toy} shows a simple example on how Eqn.\ref{propagate} plays a vital role in the saliency
propagation process.

Specifically, we choose CIE LAB color as the input feature
because distance in LAB space matches human perception well.
A color-based affinity matrix $\textbf{A}_c$ with a controlling parameter $\sigma_c$ is constructed according to Eqn.\ref{affinity},
where the feature distance $D(\textbf{f}_i,   \textbf{f}_j)= \| \textbf{c}_i-\textbf{c}_j \|_2$.
The boundary nodes are employed as the query labels
simply because regions near the image border are less likely to be salient.
However, as is shown in Fig.\ref{boundary}(b),
in some cases, the salient object appears at the border and the saliency measure
is doomed to be 0 if the salient region is chosen to be the background labels. Consequently, we
compute the color distinctiveness of each boundary node from other border regions according to Eqn.\ref{affinity},
drop the top 30\% with high color difference empirically, and thus create the set of selected boundary labels %, denoted as
$B'$.
We can also observe from Fig.\ref{boundary}(c) that the geodesic constraint with selected boundary labels facilitates
 saliency accuracy in such scenarios by strengthening the connection among boundary regions.
%\vspace{-.3cm}

%\vspace{-.6cm}

Alg.\ref{alg1} summarizes the inner label propagation via boundary nodes.
The convergence of the similarity measure $\textbf{V}$ is ensured by checking
whether its average variance in
the last 50 iterations (\emph{i.e.}, $const=49$) is below a threshold.
$\verb"sp2map"(\cdot)$ means mapping the saliency measures of $N$ regions into  an image-size map.
Note that  such a propagation framework is similar to that in \cite{Zhu03semi-supervisedlearning}.
However, we find the ranking results obtained from a closed-form solution
less encouraging than those of ours due to  different constructions of the affinity matrix.

\begin{figure*}[t]
  \centering
  \includegraphics[width=15cm]{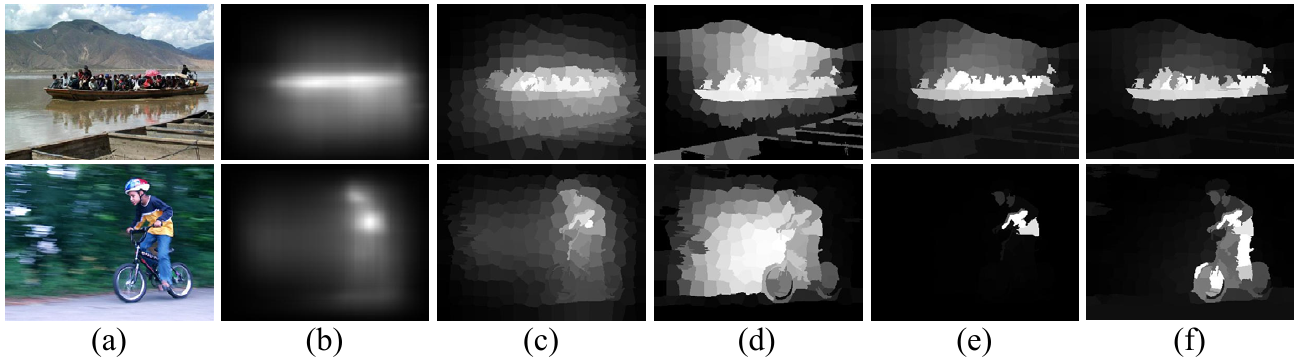}
  %\vspace{-.2cm}
  \caption{  Objectness integration. (a) Input image; (b) Pixel-level objectness map;
  (c) Region-level objectness map;
%  (d) Inner propagation via boundary labels;
%  (e) Inner propagation via objectness labels;
  (d) Inner boundary propagation;
  (e) Inner objectness propagation;
  (f) Inter propagation via boundary and objectness labels.
  }\label{obj_inte}
  %\vspace{-0.6cm}
\end{figure*}

In most cases, the inner propagation with help of the boundary labels works well whereas in some complex scenes,
as is shown in Fig.\ref{obj_inte}(d),  depending on the boundary prior alone
might lead to high saliency assignment to the background regions. It naturally suggests us to use
 some foreground prior to improve the results further.
%\vspace{-.5cm}
% Alg.1, similarity ranking
\begin{algorithm}
\renewcommand{\algorithmicrequire}{\textbf{Input:}} % Use Input in the format of Algorithm
\renewcommand{\algorithmicensure}{\textbf{Output:}}
\caption{Inner Label Propagation via Boundary Nodes}\label{alg1}
\begin{algorithmic}[1]
\Require
\Statex The $N\times N $ row-wise normalized color affinity matrix $\textbf{A}_c$.
\Statex The set of selected boundary labels $B'$ and the set of unlabelled nodes $U = \{ R \backslash B' \} $.
%\smallskip
\State $t = 0$
\State Initialize,
set $V_t(r_i) = 1 $ for  $r_i \in B'$
and $V_t(r_i) = 0$ for %those in %$U=R-B'$
$r_i \in U $%\comment{dfdf}

    \While{$check>thres$}
     \For{$r_i \in U$}
    \State $V_{t+1}(r_i)=\sum_{j=1}^{N}a_{ij}V_t(r_j)$
    \EndFor
    \State $t = t+1$
    \State $check = \verb"var"(\textbf{V}_{t},\textbf{V}_{t-1},\dots, \textbf{V}_{t-const})$
    \EndWhile
    \State $\textbf{S}^B= \verb"ones"(N) -\verb"normalize"(\textbf{V}_t)$
    \State $S^B(r_i)=\verb"sp2map"(\textbf{S}^B)$    %\Comment{$\textsf{sp2map}(\cdot)$ assigns saliency of regions to the image.}%\Comment{dfd$\verb"sp2map"(\cdot)$}
\smallskip
\Ensure
\Statex The regional  map $S^B(r_i)$ from background labels.
\end{algorithmic}
\end{algorithm}

%\subsubsection{Subsubsection Heading Here}
%Subsubsection text here.

\subsection{Objectness Labels as Foreground Prior}\label{sec:object}
%Due to the limited knowledge of boundary labels in some cases, we employ an objectness measure to specifically target the desired object and thus form a label set of the foreground.

Alexe \emph{et al.}  \cite{Objectness}  propose a novel method based on low-level cues to
compute an objectness score for any given image window,
which indicates the  likelihood of the window containing an object.
Several useful priors are exploited and combined in a Bayesian framework, including multi-scale saliency (MS), color-contrast
    (CC), edge density (ED),
superpixel straddling (SS) and location plus size (LS).
The results show high performance on the PASCAL VOC 07 dataset.

\begin{itemize}
%\vspace{-.1cm}
  \item MS, proposed by \cite{07cvpr/hou_SpectralResidual}, measures the uniqueness of objects according to the spectral residual of the image's FFT.
  \item CC,    similar as in \cite{11pami/Liu_Learning}, considers the distinct appearance of objects via a center-surround histogram of color distribution.
  \item ED and SS capture the closed boundary of objects. The former computes the density of edges near window borders while the latter calculates how intact
  superpixels are inside a window.
  \item LS exploits the likeliness of a window to cover an object based on its size and location using kernel density estimation.
\end{itemize}
%\vspace{-.1cm}

In practice, we find the first three cues more important while the last two more trivial.
Big and homogeneous superpixels are generated by \cite{pff} in \cite{Objectness} whereas small, tiny, and compact superpixels are created by the SLIC algorithm, making SS incompatible in our work.
Furthermore, LS measures the size and location of windows without taking into consideration the intrinsic features
of images and often dominates the final integrated objectness score due to different image benchmarks.
To this end, we only utilize MS, CC and ED since  cues are combined independently
in a naive Bayes model. The rest of the parameters in the objectness measure are set to be default as in \cite{Objectness}.

Let $P_m$ be a probability score of the $m$-th sampling window,
the pixel-level objectness map $\mathcal{O}(p)$ is obtained through overlapping
scores multiplied by the Gaussian smoothing kernel of
all sampling windows:
\begin{equation}
\mathcal{O}(p) = \sum_{m=1}^{M}P_m \cdot \exp \bigg[-\bigg(\frac{(x_p-x_m^c)^2}{2\sigma_x^2} + \frac{(y_p-y_m^c)^2}{2\sigma_y^2}\bigg)\bigg] \label{pixel-wise}
\end{equation}
where $M=1000$ is the number of sampling windows, $x_p, y_p, x_m^c,y_m^c$ denote the coordinates of pixel $p$
and the center coordinates of window $m$ respectively. We set $\sigma_x=0.25W$
and $\sigma_y=0.25H$, where $W$ is the width and $H$ the height of an image.
The region-level objectness map $\mathcal{O}(r_i)$
is the average of pixels' objectness values within a region:
\begin{equation}
  \mathcal{O}(r_i)= \frac{1}{n_i}\sum_{p\in r_i}\mathcal{O}(p)\label{regional}
\end{equation}
where $n_i$ indicates the number of pixels in region $r_i$.
%\vspace{-.3cm}
%\vspace{-.7cm}

The integration of objectness labels is illustrated in Fig.\ref{obj_inte}. By introducing only three cues of the objectness measure and a Gaussian kernel refinement,
the pixel-level map in Fig.\ref{obj_inte}(b) can better capture and highlight the focus of a salient object.
The region-level
objectness map in Fig.\ref{obj_inte}(c) is obtained similarly as one of the three saliency maps in \cite{UFO13}.
A simple average of pixels' scores within a region leads to mid-value saliency in vast background areas since
the pixel-level map
%\vspace{-.3cm}
%
from which the region-level map is generated is ambiguous around the salient object in the first place.

Based on the fact that high values of region-level objectness score calculated by Eqn.\ref{regional} can better indicate
  foreground areas, the set of objectness labels $O$ is created from superpixels whose
region-level objectness $\mathcal{O}(r_i)$ is no less than the objectness criterion $\gamma_1$.
Fig.\ref{obj_inte}(e) displays the saliency maps by the inner label propagation via objectness labels alone.
We observe that under the objectness mechanism, the top image effectively inhibits high
values of the background saliency while the bottom image only detects the kid's orange shirt  due to
a limited number of label hints from set $O$.
This indicates that a  complementary combination of the boundary and objectness labels could be a better choice.

\subsection{Inter Propagation via Co-transduction}\label{sec:co-trans}

Recently, Bai \emph{et al.} \cite{CoTrans} propose a similarity ranking algorithm by fusing different affinity measures for robust shape retrieval
under a semi-supervised learning framework. Inspired by such an idea, we devise a new co-transduction algorithm for saliency detection, which
%to alternate labels across different sets to refine the performance of label propagation.
uses one label set to pull out
confident data and add additional labels as new hints to the other label set.
The inter label propagation algorithm is summarized in Alg.\ref{alg2}.
Besides different application areas, our algorithm differentiates from
the original work \cite{CoTrans} in the following three ways:

%\vspace{-.6cm}
% Alg.2, co-transduction
\begin{center}
\begin{algorithm}
\renewcommand{\algorithmicrequire}{\textbf{Input:}} % Use Input in the format of Algorithm
\renewcommand{\algorithmicensure}{\textbf{Output:}}
\caption{Inter Label Propagation via Boundary and Objectness Nodes}\label{alg2}
\begin{algorithmic}[1]

\Require
\Statex The $N\times N $ row-wise normalized color affinity matrix $\textbf{A}_c$.
\Statex The set of selected boundary labels $B'$ and the set of objectness labels $O$.
\smallskip
\State $t=0$
\State Initialise, $\textbf{V}_t^{B}=\textbf{0},\textbf{V}_t^{O}=\textbf{0}$

    \While{$check^{B},check^{O}>thres$}
    \smallskip
    \State Set $V_{t}^{B}(r_i)=1$ for $r_i \in B'$, $V_{t}^{O}(r_i)=1$ for $r_i \in O$
    \State Create unlabelled sets $U_1$ and $ U_2$ such that $U_1=\{R \backslash B'\}, U_2=\{R\backslash O\}$
    \smallskip
    \For {$r_i \in U_1, r_i \in U_2$}
    \State $V_{t+1}^{B}(r_i)=\sum_{j=1}^{N}a_{ij}V_t^{B}(r_j)$
     \State $ V_{t+1}^{O}(r_i)=\sum_{j=1}^{N}a_{ij}V_t^{O}(r_j)$
    \EndFor

    \State $t=t+1$
    \smallskip
    \State $check^{B} = \verb"var"(\textbf{V}_{t}^{B},%\textbf{V}_{t-1}^{B},
    \dots, \textbf{V}_{t-const}^{B})$

    \State $check^{O} = \verb"var"(\textbf{V}_{t}^{O},%\textbf{V}_{t-1}^{O},
    \dots, \textbf{V}_{t-const}^{O})$
    \smallskip
    \State $temp1=\verb"sort"(\textbf{V}_{t}^{B},\verb"`ascend'")$
    \State $temp2=\verb"sort"(\textbf{V}_{t}^{O},\verb"`ascend'")$
%    \State $\textbf{V}_{t+1}^{B}=\verb"LP"(\textbf{V}_{t}^{B}, \textbf{A}, B)$,
%    $\textbf{V}_{t+1}^{O}=\verb"LP"(\textbf{V}_{t}^{O}, \textbf{A}, O)$
\smallskip
    \State $L^B=temp1(1:p_1)$, $L^O=temp2(1:p_2)$
    \State $B'= B' \cap L^O$, $O= O \cap L^B$
   \smallskip

    \EndWhile
     \State $\textbf{S}^B= \verb"ones"(N) -\verb"normalize"(\textbf{V}_t^B)$

     \State $\textbf{S}^O= \verb"normalize"(\textbf{V}_t^O)$,
     %$\textbf{S}=\verb"norm"(\frac{\textbf{S}^B+\textbf{S}^O}{2})$

\State $\textbf{S}^{C}=\verb"normalize"(\alpha\textbf{S}^B+\beta\textbf{S}^O)$
\State $S^{C}(r_i)=\verb"sp2map"(\textbf{S}^{C})$
\smallskip
\Ensure
\Statex The combined regional saliency map    $S^{C}(r_i)$.
\end{algorithmic}
\end{algorithm}
\end{center}
%\vspace{-.4cm}
%
%\vspace{-.5cm}

\begin{figure}
  \centering
  \includegraphics[width=6cm]{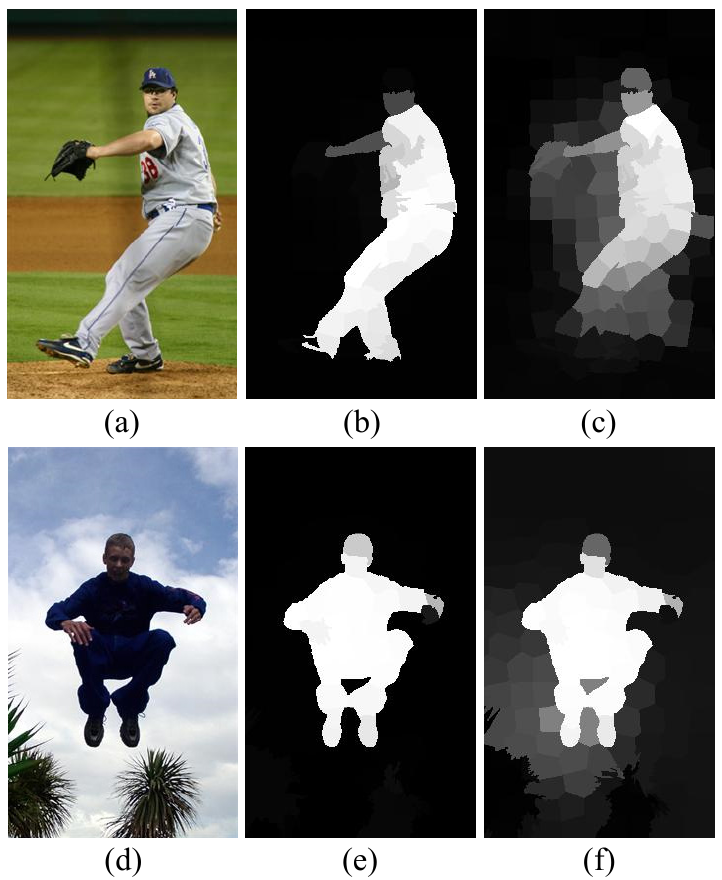}
  \vspace{-.2cm}
  \caption{ Effects of co-transduction algorithm. (a) Input image; (b) Only color affinity matrix;
  (c) Both color and structure affinity matrix; (d) Input image;
  (e) Saliency map by Alg.\ref{alg1};
  (f) Saliency map by Alg.\ref{alg2}.
  }\label{co_trans}
\end{figure}

First, instead of fusing two
different similarity matrices, we %find the structural description to be ineffective and
construct
the same matrix $\textbf{A}_c$ for both label sets (through line 7 to 8). Fusing two affinity matrices is investigated
and an orientation-magnitude (OM) descriptor \cite{PISA13} is extracted
to capture the structural characteristic of images. We compute a structure-based affinity matrix $\textbf{A}_s$ % is generated
according to Eqn.\ref{affinity}, where $D(\textbf{f}_i, \textbf{f}_j)=\chi^2(h_{OM}(i), h_{OM}(j))$ and $\sigma_{s}^{2}=0.1$.
As shown in Fig.\ref{co_trans}(a)-(c), the saliency map using one color affinity matrix outperforms that of using two matrices.
The information from structure description seems to be redundant since
%images are divided into regions based on colors and edges. That is to say,
the color affinity matrix $\textbf{A}_c$ already includes
knowledge of textual distinctiveness    at the borders of  each region.

Second, we emphasize more on the difference between boundary and objectness labels during the propagation
%while take less consideration on different affinity matrices.
whereas the same $p$ labels are switched within each label set in \cite{CoTrans}.
During each iteration in Alg.\ref{alg2} (through line 11  to 16), $p_1$ superpixels which are most different from the  boundary labels
are picked out and added to the objectness set and the update of the boundary set is
similarly achieved with a different superpixel number $p_2$.
We set $p_1, p_2$ to be $p_1 \ll p_2$ because the background regions %often exceeds much more
often significantly outnumber  the foreground ones.

Third, we observe that the ranking values in the first few recursions to be highly noisy and inaccurate. Therefore, unlike the practice of \cite{CoTrans} that averages all the similarity measures in each iteration, the final saliency measure is computed as a linear combination of the resultant
$\textbf{S}^B$ and $\textbf{S}^O$ in the last iteration from boundary and objectness labels, respectively (through line 18 to 21).

From Fig.\ref{obj_inte}(d)-(f) we can see that such a  co-transduction algorithm outperforms
the inner label propagation via boundary or objectness nodes alone.
The failure images in the inner boundary propagation are often cases where there are vast areas with high-value saliency assignment to regions around the salient object.
The reason ascends from the resemblance of appearance between salient and non-salient nodes, as well as the spatial discontinuity from the boundary labels
%to propagate its saliency to the center boundary-like regions.
to the center regions which prevent labels from being propagated to the background regions around the image center.
The inter propagation algorithm strengthens the connection of salient regions by employing objectness labels and distinguishes the foreground better from the background by enlarging the set of boundary labels from objectness cues, thus best leveraging the complimentary information of both label sets.
Some may argue that the graph cuts optimization method also performs well in saliency detection \cite{CBS11}, but the co-transduction algorithm is designed to obtain continuous saliency assignment while the former aims at solving binary MRF problems.
%\vspace{-.5cm}
%\vspace{-.7cm}

In some cases, as shown in Fig.\ref{co_trans}(d)-(f), the inner propagation via boundary labels alone has better saliency maps than a combination of boundary
and objectness labels, which results from the slight disturbance of objectness measures near the salient object.
To this end, we propose a compactness score to evaluate the quality of the regional saliency map $S^B(r_i)$
generated by Alg.\ref{alg1}:
%\vspace{-.1cm}
\begin{equation}\label{compactness}
  C(S) = \sum_{b=1}^{10}w(b)\cdot h^{S}(b)
  \vspace{-.1cm}
\end{equation}
where $b$ denotes each quantisation of the resultant saliency map, $h^{S}(b)$ indicates a 10-bin histogram distribution of the map and $w(b)$ indicates the weight upon each bin.
%
%The quality of vast areas of mid-value saliency assignment to the map (see Fig.\ref{obj_inte}(d)) is much inferior
%to that of clear high or low-value saliency assignment. Therefore,
%
Based on the aforementioned characteristic of the failure saliency maps in the inner boundary propagation,
we take a triangle  form of the weight term,
\emph{i.e.}, $w(b)=\min(b,(11-b))$.
%
%Then a compactness criterion $\gamma_2$ is introduced to decide whether the saliency map, according to Enq.\ref{compactness}, needs a further update.
Only the saliency maps with score lower than a compactness criterion $\gamma_2$ will be updated by the inter propagation via a
co-transduction algorithm. Such a scheme not only ensures high quality of the saliency maps, but also improves the computational efficiency.

\subsection{Pixel-level Saliency Coherence}\label{sec:coherence}

Finally, in order to eliminate the segmentation errors of the SLIC algorithm,
we define the pixel-level saliency %$S(p)$
as a weighted linear combination
of the regional saliency, $S^{B}(r_i)$ or $S^{C}(r_i)$ %\footnote{   Here $S^{B}$ or $S^{C}$ is the straightforward region-level result descending from Alg.\ref{alg1} or Alg.\ref{alg2}. }
, of its surrounding superpixels:
\begin{equation}\label{}
  S(p)=\sum_{i=1}^{G}%\frac{1}{Q_i}
  \exp\big(- (k_1 \| \textbf{c}_p-\textbf{c}_i \| + k_2 \| \textbf{z}_p-\textbf{z}_i \|)  \big)S^{B/C}(r_i)
\end{equation}
where $\textbf{c}_p,\textbf{c}_i, \textbf{z}_p, \textbf{z}_i$ are the color and coordinate vectors of a region or a pixel, $G$ denotes the number of direct neighbors of region $r_i$, and $S^{B}$ or $S^{C}$ indicates the straightforward region-level result descending from Alg.\ref{alg1} or Alg.\ref{alg2}. %, and $Q_i$ indicates the partition function.
By choosing a Gaussian weight, we ensure the up-sampling process is both local and color sensitive. Here $k_1$ and $k_2$
are parameters controlling the sensitivity to color and position, where
$k_1=0.2, k_2=0.01$ is found to work well in practice.

\begin{figure*}
  \centering
  \includegraphics[width=\textwidth]{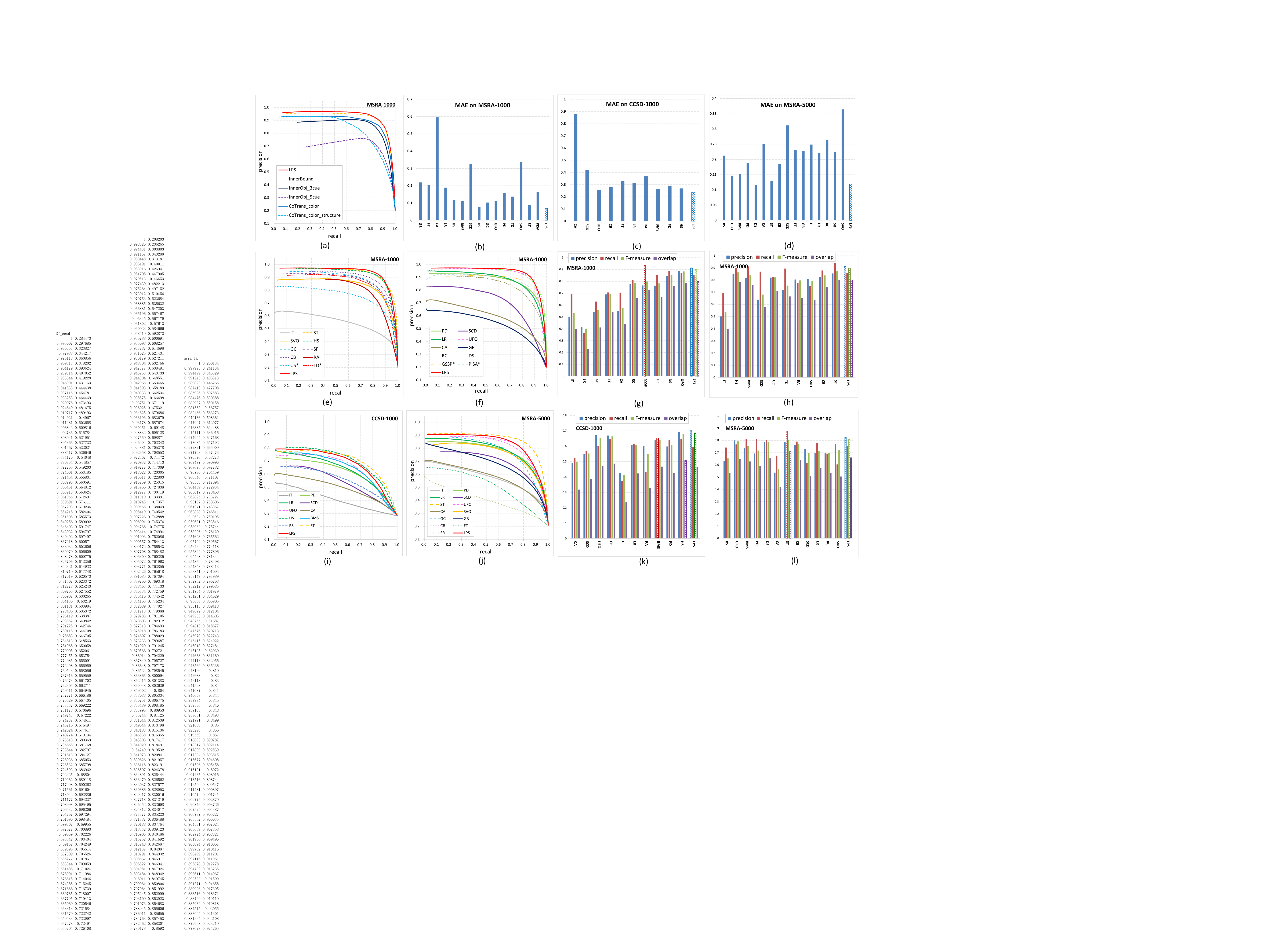}
  \vspace{-.4cm}
  \caption{ Quantitative  results. (a) Individual component analysis on MSRA-1000. Note that
  `CoTrans\_$*$' means implementing Alg.\ref{alg2} for every image;
  (b)-(d) \emph{MAE} metric on MSRA-1000, CCSD-1000, MSRA-5000; (e)-(l) Performance comparison on MSRA-1000, CCSD-1000 and MSRA-5000 respectively.
  Bars with oblique lines denote the highest score in the corresponding metric. Methods followed by an asterisk (*) denote they are only compared in those datasets.
  }\label{pr_1_new}
\end{figure*}

\section{Experimental Results}
We evaluate the proposed method on five typical datasets.
The first MSRA-1000, which is a subset of MSRA-5000,
is a widely used dataset where almost every method has been tested by comparing to the
accurate human-labelled masks  provided in \cite{09cvpr/Achanta_FTSaliency}.
The second CCSD-1000 \cite{Hierarchy} contains more salient objects under complex scenes
and some images come from the challenging Berkeley-300 dataset \cite{GS12}.
The third MSRA-5000 \cite{11pami/Liu_Learning} includes a more comprehensive source of images with
accurate masks recently released by \cite{DRFI13}.
The fourth THU-10,000 is the largest dataset %\footnote{\href{http://mmcheng.net/SalObj/}{http://mmcheng.net/SalObj/}}
in the saliency community so far where
10,000 images are randomly chosen from the MSRA database and the Internet with pixel-level labeling.
The last PASCAL-S \cite{Li_2014_CVPR} ascends from the validation set of PASCAL VOC 2010 segmentation challenge.
It contains 850 natural images where in most cases multiple objects of varying size, shape, color, etc., are surrounded by complex scenes.
Unlike the traditional benchmarks, the PASCAL-S is believed to
eliminate the dataset design bias.

The proposed LPS algorithm is compared with both the classic and  newest state-of-the-arts:
IT\cite{IT98},
%MZ\cite{03ACMMM/Ma_Contrast-based},
GB\cite{07ANIPS/harel_graph},
SR\cite{07cvpr/hou_SpectralResidual},
%AC\cite{08cvs/achanta_salient},
LC\cite{06LC},
FT\cite{09cvpr/Achanta_FTSaliency},
CA\cite{10cvpr/goferman_context},
RA\cite{Finland10},
CB\cite{CBS11},
SVO\cite{SVO11},
HC\cite{RC11},
BS\cite{DUT11},
SF\cite{SF12},
LR\cite{Northwestern12},
GSSP\cite{GS12},
%, and the most recent ones (
MK\cite{Markov_dlut}, DS\cite{PCA_dlut},
GC\cite{GC13}, PD\cite{PatchDistinct13},
MR\cite{Manifold}, BMS\cite{Boolean},
HS\cite{Hierarchy}, US\cite{US13},
UFO\cite{UFO13}, TD\cite{TD13},
PISA\cite{PISA13},
HPS\cite{Hyper13},
ST\cite{ST},
SCD\cite{SCD}.
%DRFI\cite{DRFI13}
%).
%Literatures of the comparison methods is notated in the corresponding figures.
To evaluate
these methods, we either use results provided by authors
or run their implementations based on the available codes or softwares.

\subsection{Parameters and Evaluation Metrics}

\subsubsection{Implementation Details}

We set the        control of color distance $\sigma_{c}$ in Eqn.\ref{affinity}
to be $\sigma_{c}^{2}=0.1$%\footnote{ This setting is very common in the saliency community\cite{Manifold,UFO13,Hyper13}.}
, the number of switching labels from background and objectness labels
in Alg.\ref{alg2} to be $p_1=2, p_2=150$, respectively.
The objectness criterion associated with Eqn.\ref{regional} is chosen to be $\gamma_1=0.8$ and
the compactness criterion with Eqn.\ref{compactness} is fixed at $\gamma_2=1.6$.
%
%A slight lean to boundary labels in the combination coefficients, \emph{i.e.}, $\alpha = 0.6, \beta=0.4$,
%would favor the results.
Parameters are empirically selected (see Tab.\ref{param_cmp} and Sec.\ref{para_sec}) and universally used for all images.

\subsubsection{Fixed Threshold}
In the first experiment we compare binary masks for every threshold
in the range $[0,\dots,255]$ and calculate the precision and recall rate.
Precision
corresponds to the percentage of salient pixels correctly
assigned, while recall corresponds to the fraction of
detected salient pixels in relation to the number
of salient pixels in ground truth maps.

\subsubsection{Adaptive Threshold}
In the second experiment we employ the saliency-map-dependent
threshold proposed by \cite{09cvpr/Achanta_FTSaliency} and define it as proportional to the mean saliency of a map:
\begin{equation}\label{}
  T_a = \frac{k}{W\times H}\sum_{x=1}^{W}\sum_{y=1}^{H}S(x,y)
\end{equation}
%$T_a = \frac{k}{W\times H}\sum_{x=1}^{W}\sum_{y=1}^{H}S(x,y)$
where
%$W$ and $H$ are the width and height of the saliency map $S$ and we typically choose $k=1.5$.
$k$ is typically chosen to be 1.5 \cite{UFO13}.
Then a weighted harmonic mean measure between precision and recall, \emph{i.e.}, F-measure, is introduced by
\begin{equation}\label{}
  F_{\beta}=\frac{(1+\beta^2)Precision \times Recall}{\beta^2 \times Precision + Recall}
\end{equation}
%$F_{\beta}=\frac{(1+\beta^2)Precision \times Recall}{\beta^2 \times Precision + Recall}$
where we set  $\beta^2=0.3$  to emphasize precision \cite{09cvpr/Achanta_FTSaliency}.
As we can see later, one method cannot have in all the highest precision, recall and F-measure as the former two are mutually exclusive
and the F-measure is a complementary metric to balance them.

Furthermore, the overlap rate $R_o$ defined by the PASCAL VOC criterion
(\textit{i.e.,} intersection over union) is used to comprehensively leverage precision and recall under the
adaptive-threshold framework.

\subsubsection{Mean Absolute Error}
 In the third experiment we introduce
the mean absolute error (MAE) between the continuous saliency map $S$ and
the binary mask of ground truth $GT$:
%\vspace{-.3cm}
\begin{equation}\label{}
  MAE=\frac{1}{W\times H}\sum_{x=1}^{W}\sum_{y=1}^{H}| S(x,y)-GT(x,y) |.
  %\vspace{-.2cm}
\end{equation}
The metric takes the true negative saliency assignments into account whereas  the precision
and recall favor the successfully assigned saliency to the salient pixels \cite{GC13}.
Moreover, the quality of the weighted continuous saliency maps may
be of higher importance than the binary masks in some cases \cite{SF12}.

%\subsubsection{Revised F-measure}
%
%
%In the fourth experiment, we evaluate our model
%on the PASCAL-S dataset with a revised F-measure \cite{Margolin_2014_CVPR}.
%This metric overcomes some inherent flaws in the traditional precision and recall metrics by directly
%comparing the non-binary maps with ground truth maps, thus providing a more objective evaluation among  algorithms.
%For the detailed definitions, we refer readers to \cite{Margolin_2014_CVPR}.

\subsection{Quantitative Comparison}

\subsubsection{Individual Component Analysis}

In order to demonstrate the effects of separate components and their combinations
in our approach, we plot the precision-recall curves in Fig.\ref{pr_1_new}(a).

First, we see that the refined co-transduction algorithm (LPS) with a compactness selection mechanism outperforms the  inner propagation  via boundary labels or objectness ones alone. Second, the precision rate under the two-feature-matrices framework in the co-transduction (blue dashed line)
goes down sharply at high recall, which indicates the structure descriptor cannot inhibit the background regions. Third,
the  take-all-cue scheme from \cite{Objectness} fails to achieve high precision especially at higher thresholds, which verifies
our explanations in Sec.\ref{sec:object} to take only three cues.
%
%Note that there is a slight rise in the P-R curves in the inner objectness propagation. This is because
%inaccurate objectness labels are chosen from the non-salient regions, leading to the imprecision at higher recall.
%\vspace{-0.3cm}
Note that in the inner objectness propagation, the precision at lower recall is even slightly worse because of the inaccurate objectness labels chosen from the non-salient regions.

\begin{table*}
 \renewcommand{\arraystretch}{1.5}
    \centering
  \caption{ Execution time comparison in second unit per image on the MSRA-1000 dataset.
   All codes are downloaded from the authors' website  and run unchanged in MATLAB 2013a with some methods' C++ MEX implementation.}\label{tab:Time}
  %\smallskip
    \begin{tabular}{l|c|c|c|c|c|c|c|c|c|c|c|c|c } \hline
      Method   & Alg.\ref{alg1}  &  Alg.\ref{alg2} & LPS  & UFO\cite{UFO13} & SVO\cite{SVO11} &CB\cite{CBS11} & PD\cite{PatchDistinct13} &HPS \cite{Hyper13} &LR\cite{Northwestern12} &CA\cite{10cvpr/goferman_context}  & DS\cite{PCA_dlut} & PD\cite{PatchDistinct13} & HPS\cite{Hyper13}\\ \hline
      Time(s) & 0.87 & 9.56 & 2.45 & 18.73 & 40.33 &1.18& 3.64 &5.02& 11.92 &36.05 & 0.84 & 19.45 & 3.16\\ \hline
    \end{tabular}
\end{table*}

\begin{table*}

%\small{
 \renewcommand{\arraystretch}{1.5}
\caption{ Parameter Selection and Model Robustness. We test different parameters on three benchmarks in terms of F-measure (higher is better) and MAE (lower is better).
The best parameters are written in \textbf{bold}, which are our model's default settings. \textbf{\textcolor[rgb]{1.00,0.00,0.00}{Red}} and \textcolor[rgb]{0.00,0.00,1.00}{\textbf{blue}} numbers in bold represent the best and better performance in each evaluation category.
%The best three results are highlighted
%in \textcolor{red}{red}, \textcolor{blue}{blue}  and  \textcolor{green}{green}, respectively.
} \label{param_cmp}

\begin{center}
\begin{tabular}{c||c|c|c|c|c|c|c|c|c|c|c|c |c}
\hline
%\textcolor{green}{0.67}
\multirow{2}{*}{Dataset}     &  \multirow{2}{*}{Metric} 	 & \multicolumn{3}{c|}{$\#$ of switching labels $p_1, p_2$}    & \multicolumn{5}{c|}{The objectness criterion $\gamma_1$}   & \multicolumn{4}{c}{The compactness criterion $\gamma_2$}  \\\cline{3-14}
    &   &   \textbf{2, 150}&     $5, 175$ & $10, 200$  &   $0.4$ &$0.6$ &\textbf{0.8} &$1.0$ &$1.2$ &$1.4$ &\textbf{1.6} &   $1.8$   &  $2.0$ \\ \hline
\multirow{2}{*}{MSRA-1000} & F-measure 	& \textbf{\textcolor[rgb]{1.00,0.00,0.00}{0.90}} & \textcolor[rgb]{0.00,0.00,1.00}{\textbf{0.83}} & 0.52  &  0.72   & 0.84  & \textbf{\textcolor[rgb]{1.00,0.00,0.00}{0.90}}  & \textcolor[rgb]{0.00,0.00,1.00}{\textbf{0.87}} &  0.80 &  0.83 & \textcolor[rgb]{1.00,0.00,0.00}{\textbf{0.90}} & \textcolor[rgb]{0.00,0.00,1.00}{\textbf{0.87}}   &  0.79 \\\cline{2-14}
	& MAE		& \textcolor[rgb]{1.00,0.00,0.00}{\textbf{0.07}} & \textcolor[rgb]{0.00,0.00,1.00}{\textbf{0.13}} & 0.31  & 0.20 &0.11  & \textcolor[rgb]{1.00,0.00,0.00}{\textbf{0.07}}   & \textcolor[rgb]{0.00,0.00,1.00}{\textbf{0.09}} & 0.13 & 0.12  & \textbf{\textcolor[rgb]{1.00,0.00,0.00}{ 0.07}} & \textcolor[rgb]{0.00,0.07,1.00}{\textbf{0.09}} &0.16  \\\hline
\multirow{2}{*}{CCSD-1000}& F-measure 	& \textcolor[rgb]{1.00,0.00,0.00}{\textbf{0.68}} &  \textbf{\textcolor[rgb]{0.00,0.07,1.00}{0.56}} &  0.55 & 0.60 & \textbf{\textcolor[rgb]{1.00,0.00,0.00}{0.683}} & \textcolor[rgb]{0.00,0.07,1.00}{\textbf{0.681} }&  0.64&  0.58 & 0.63 &   \textbf{\textcolor[rgb]{1.00,0.00,0.00}{0.68}} &\textcolor[rgb]{0.00,0.00,1.00}{\textbf{0.66}} & 0.60 \\\cline{2-14}
					& MAE		& \textcolor[rgb]{1.00,0.00,0.00}{\textbf{0.23}} &  \textbf{\textcolor[rgb]{0.00,0.00,1.00}{0.32}} &   0.35 & 0.33 &  \textbf{\textcolor[rgb]{1.00,0.00,0.00}{0.21}}&  \textcolor[rgb]{0.00,0.07,1.00}{\textbf{0.23}}  & 0.28 &  0.35&  0.27 &  \textcolor[rgb]{1.00,0.00,0.00}{\textbf{ 0.23 }}& \textcolor[rgb]{0.00,0.00,1.00}{\textbf{0.26}} & 0.32 \\\hline
\multirow{2}{*}{MSRA-5000}& F-measure 	& \textcolor[rgb]{1.00,0.00,0.00}{\textbf{0.81}}  &  \textbf{\textcolor[rgb]{0.00,0.07,1.00}{0.77}} & 0.63 &0.75 & 0.79  &\textcolor[rgb]{0.00,0.00,1.00}{\textbf{0.81}} &  \textcolor[rgb]{1.00,0.00,0.00}{\textbf{0.83}} &   0.78 & 0.77 & \textcolor[rgb]{1.00,0.00,0.00}{\textbf{0.813}} &0.78  & \textcolor[rgb]{0.00,0.07,1.00}{\textbf{0.811}} \\\cline{2-14}
					& MAE		& \textbf{\textcolor[rgb]{1.00,0.00,0.00}{0.12}} &   \textcolor[rgb]{0.00,0.07,1.00}{\textbf{0.25}} & 0.30 & 0.20  & 0.16 & \textbf{\textcolor[rgb]{0.00,0.07,1.00}{0.12}}  & \textbf{\textcolor[rgb]{1.00,0.00,0.00}{0.11}} &  0.16 & 0.25 &  \textcolor[rgb]{1.00,0.00,0.00}{\textbf{0.122}} &  0.15 &\textcolor[rgb]{0.00,0.07,1.00}{\textbf{0.124}}        \\\hline					
\end{tabular}

\end{center}

%}

%\vspace{-2mm}

\end{table*}

\begin{figure}
  \centering
  \includegraphics[width=8.8cm]{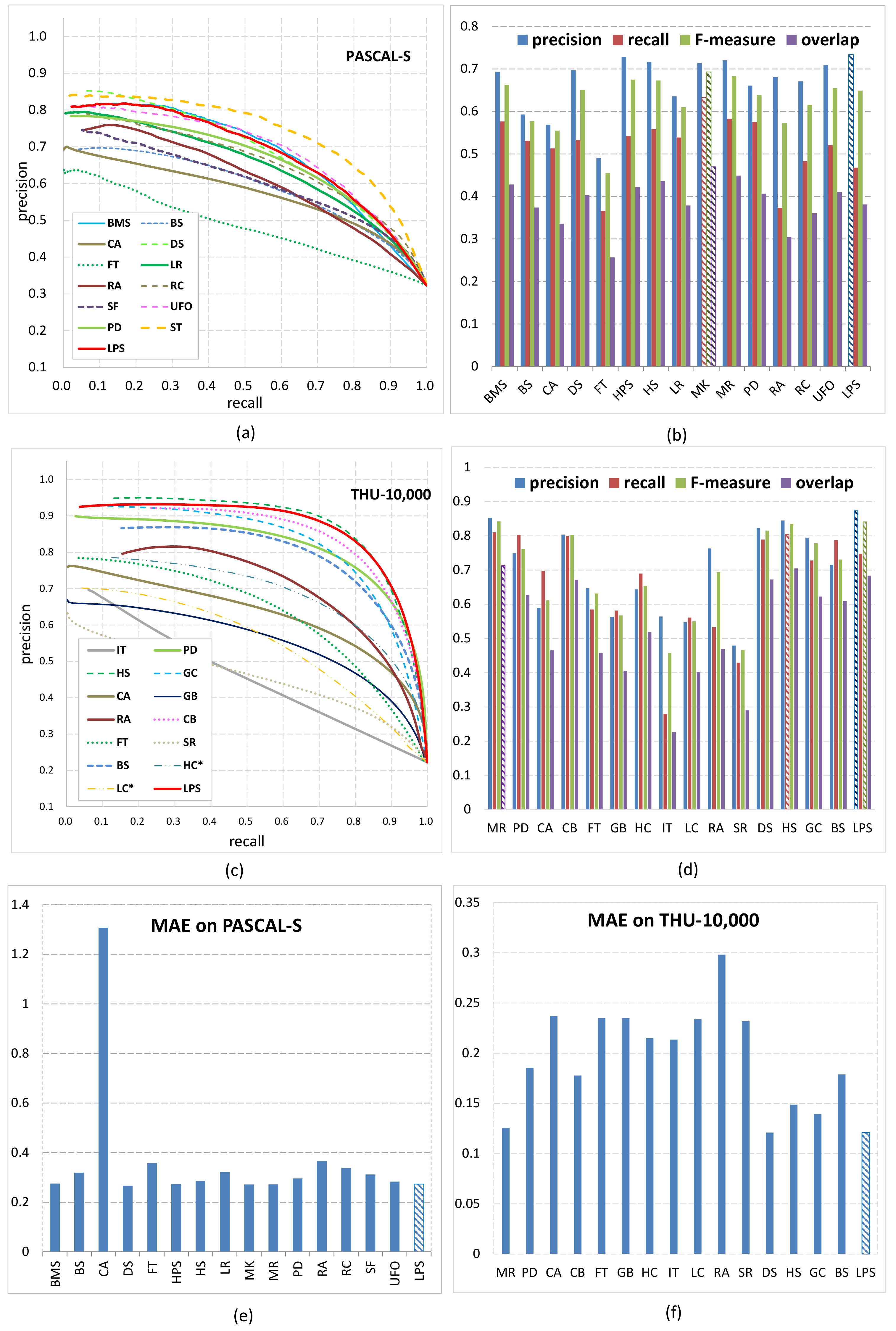}
  \vspace{-.6cm}
  \caption{ Performance of the proposed algorithm compared with previous methods on  the PASCAL-S (a),(b),(e) and
  THU-10,000 (c),(d),(f).
  Bars with oblique lines denote the highest score in the corresponding metric. Methods followed by an asterisk (*) denote they are only compared in those datasets.
  }\label{pr_2_new}
\end{figure}

\smallskip
\subsubsection{Mean Absolute Error}

Fig.\ref{pr_1_new}(b)-(d) shows the MAE metric of LPS and other methods on MSRA-1000, CCSD-1000 and MSRA-5000.
Considering the recent and well-performed methods, such as
DS13\cite{PCA_dlut},  GC13\cite{GC13}, BMS13\cite{Boolean}, TD13\cite{TD13}, HS13\cite{Hierarchy}, PISA13\cite{PISA13},
PD13\cite{PatchDistinct13},
LPS achieves the lowest error of 0.0695, 0.2369, 0.1191 on the corresponding datasets, which indicates the resultant maps
have a high quality of highlighting salient objects while suppressing the background.
\smallskip
\subsubsection{MSRA-1000}

Fig.\ref{pr_1_new}(e)-(h) displays the P-R curves, F-measure and overlap rate on  MSRA-1000 benchmark.
On one hand, LPS achieves an average of $97\%$ precision rate covering most ranges of the recall  while models such as
DS13\cite{PCA_dlut}, UFO13\cite{UFO13}, HS13\cite{Hierarchy}, ST\cite{ST}, have similar performance competing ours
and yet lower precision at specific ranges of recall; on the other hand,
the highest precision, F-measure and overlap score of 0.91, 0.90, 0.80 is accomplished by LPS outperforming other 21 methods.
Note that due to many false positive salient detections of GSSP\cite{GS12}, their model has the highest recall value in Fig.\ref{pr_1_new}(g); however, it is more important
to have a high value of precision or F-measure in the saliency community.

\smallskip
\subsubsection{ CCSD-1000 and MSRA-5000}

The last row of Fig.\ref{pr_1_new} reports the performance comparison on these two datasets.
For the CCSD benchmark, we observe that although HS13\cite{Hierarchy} achieves better precision  curve
and higher overlap score, LPS have the highest precision of 0.705, lowest MAE error and similar F-measure.

For the MSRA-5000, compared with most methods, LPS achieves the best curve performance spanning most ranges of recall as well as  the highest precision and F-measure of 0.82, 0.81, respectively.
We observe that the ST\cite{ST} model has a competitive high value of precision in the recall range from 0.7 to 1.0 (see Fig.\ref{pr_1_new}(j)), which means they have a strong capability to suppress the image background (even assigning small saliency value is not allowed). This advantage is probably attributed to the sentimental hierarchical analysis and the multi-scale scheme in their work. As for the adaptive threshold comparison, we have reached the highest precision and similar F-measure whereas ST keeps the highest overlap and recall. At last, the lowest MAE error is accomplished by LPS on this dataset.

\smallskip

\subsubsection{PASCAL-S and THU-10,000}

Fig.\ref{pr_2_new} shows the performance comparison with other algorithms on the THU-10,000 and PASCAL-S benchmarks, in terms of a continuous-map (PR-curve) and an adaptive-threshold evaluation.
We achieve comparable performance with the best results reported so far.
Specifically, the F-measure and  precision are the highest as well as MAE the lowest on the THU dataset;
on the PASCAL-S, LPS is less inferior than MR\cite{Manifold} and MK\cite{Markov_dlut} in terms of F-measure and overlap value while we achieve the highest precision and a comparable MAE result with the best ones (HS\cite{Hierarchy}, DS\cite{PCA_dlut}). Note that the MAE of CA\cite{10cvpr/goferman_context} is quite high on every dataset because the size of their saliency maps  are much smaller than the original images\footnote{ Most saliency maps are of size $300 \times 400$ as well as the size of ground truth maps; due to the multi-scale implementation in \cite{10cvpr/goferman_context}, the larger dimension of their maps is fixed to be 250. For fair comparison, we resize their smaller results to the same size of the ground truth maps and compute the corresponding evaluation metrics.}.

\smallskip
\subsubsection{Execution Time}

Tab.\ref{tab:Time} shows the average execution time of processing one image in the MSRA-1000 dataset.
Experiments are conducted on an Intel Core i7-3770 machine, equipped with 3.40GHz dominant frequency and 32 GB RAM.
%
%By introducing the compactness criterion scheme, the efficiency of implementing LPS has increased 74\% from that
%of executing the inter propagation alone (Alg.\ref{alg2}). We observe that both UFO\cite{UFO13} and SVO\cite{SVO11},
%which utilize the objectness measure directly for every image, have suffered from poor efficiency as well as inferior P-R curves.
%
Alg.\ref{alg2} takes much longer than Alg.\ref{alg1} because the calculation of the objectness measure \cite{Objectness} is time consuming.  By introducing a selection scheme using the compactness criterion, the
computational efficiency of LPS has increased 74\%. In contrast, those methods that directly utilize the objectness measure for each single image (UFO\cite{UFO13}, SVO\cite{SVO11}) have suffered from poor efficiency as well as inferior P-R curves.

Note that some methods such as CB\cite{CBS11} and DS\cite{PCA_dlut} have faster efficiency than ours; we believe an effective parallelized acceleration using GPU implementation on the compactness calculation and the pixel-wise saliency coherence at a  pixel basis can substantially improve the computational efficiency.

%\vspace{-.9cm}
\smallskip

\subsubsection{Parameter Selection and Model Robustness}\label{para_sec}
Tab.\ref{param_cmp} shows the quantitative results using different parameter combinations. We choose the best qualified parameters in terms of F-measure and MAE on the MSRA (1000 or 5000) and CCSD datasets.
Our algorithm takes the least number of parameters in order to better generalise on different datasets.

\subsection{Visual Comparison}
\begin{figure*}
  \centering
  \begin{overpic}[width=\textwidth]{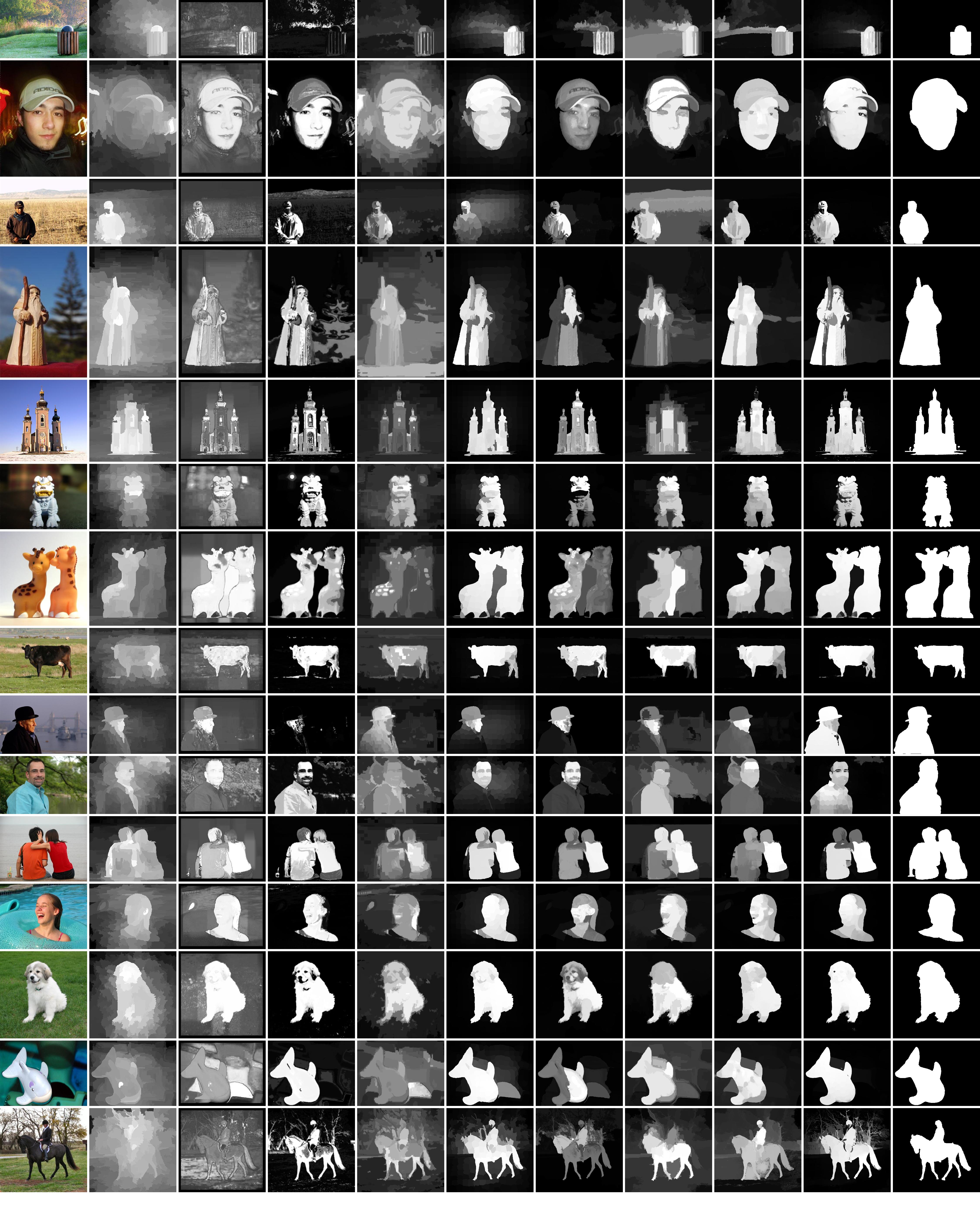}
  \put(.7, .3){\small
  ~~Source    ~~~~~SVO\cite{SVO11} ~~~~ RA\cite{Finland10} ~~~~~GC\cite{GC13} ~~~~~RC\cite{RC11} ~~~~~MK\cite{Markov_dlut} ~~~~UFO\cite{UFO13} ~~~~~~CB\cite{CBS11} ~~~~ ST\cite{ST} ~~~~~~ LPS     ~~~~~~~~ GT
  }
  \end{overpic}
  \caption{ Visual comparison of previous methods, our algorithm (LPS) and ground truth (GT) on the MSRA-5000 dataset. The last example shows a failure case where LPS overwhelmingly highlights the background around the horse due to a complex configuration of color and texture in the background.
  } \label{fig:msra_5k}
\end{figure*}

%\vspace{-0.2cm}
Several natural images with complex background are shown through Fig.\ref{fig:msra_5k} to Fig.\ref{fig:thus} for visual comparison of our method w.r.t. the most recent state-of-the-arts.
From these examples, we can see that most saliency detectors can effectively
handle cases with relatively simple background and homogenous objects, such as the third and fourth row from the bottom in Fig.\ref{fig:msra_5k}, the first three rows in Fig.\ref{fig:thus}, etc.

However, our model can tackle even more complicated scenarios, for example:
%\begin{itemize}
%  \item Cluttered background: \\
%  Row 1,2,4 in Fig.\ref{fig:cmp}; row 7, 12 in Fig.\ref{fig:thus}.
%  \smallskip
%  \item Low contrast between objects and background:\\
%  Row 4,11,13 in Fig.\ref{fig:thus}.
%  \smallskip
%  \item Heterogeneous objects: \\
%  Row 2,10,11 in Fig.\ref{fig:msra_5k}.
%  \smallskip
%  \item Multi-scale objects:\\
%  The first three rows in Fig.\ref{fig:msra_5k}.
%\end{itemize}
(a) \emph{cluttered background}: row 1,2,4 in Fig.\ref{fig:cmp}, row 7, 12 in Fig.\ref{fig:thus};
(b) \emph{low contrast between objects and background}: row 4,11,13 in Fig.\ref{fig:thus};
(c) \emph{heterogeneous objects}, row 2,10,11 in Fig.\ref{fig:msra_5k};
(d) \emph{multi-scale objects}, the first three rows in Fig.\ref{fig:msra_5k}.
More examples can be found in these figures. Due to the simple inner propagation process, our algorithm can effectively separate background labels and assign high saliency values to the dissimilar superpixels, \emph{i.e.}, the candidate salient objects. With the help of a foreground proposal scheme, \emph{i.e.}, objectness, the inter propagation can redirect the selection of foreground labels and compensate the intermediate results from the inner stage, thus detecting more accurate salient objects even from low contrast foreground and cluttered background.

%For example, the salient objects are precisely extracted in the first and sixth images whereas they appear along
%with many other false-positives in other algorithms.
%%
%LPS achieves better visual results by suppressing the background near the salient object(s).
%
%Some argue that the person's legs in the third or fifth are somewhat suppressed, this may result from the background disturbance in the boundary label set.

\subsection{Limitation and Analysis}

\begin{figure*}
  \centering
  \begin{overpic}[width=\textwidth]{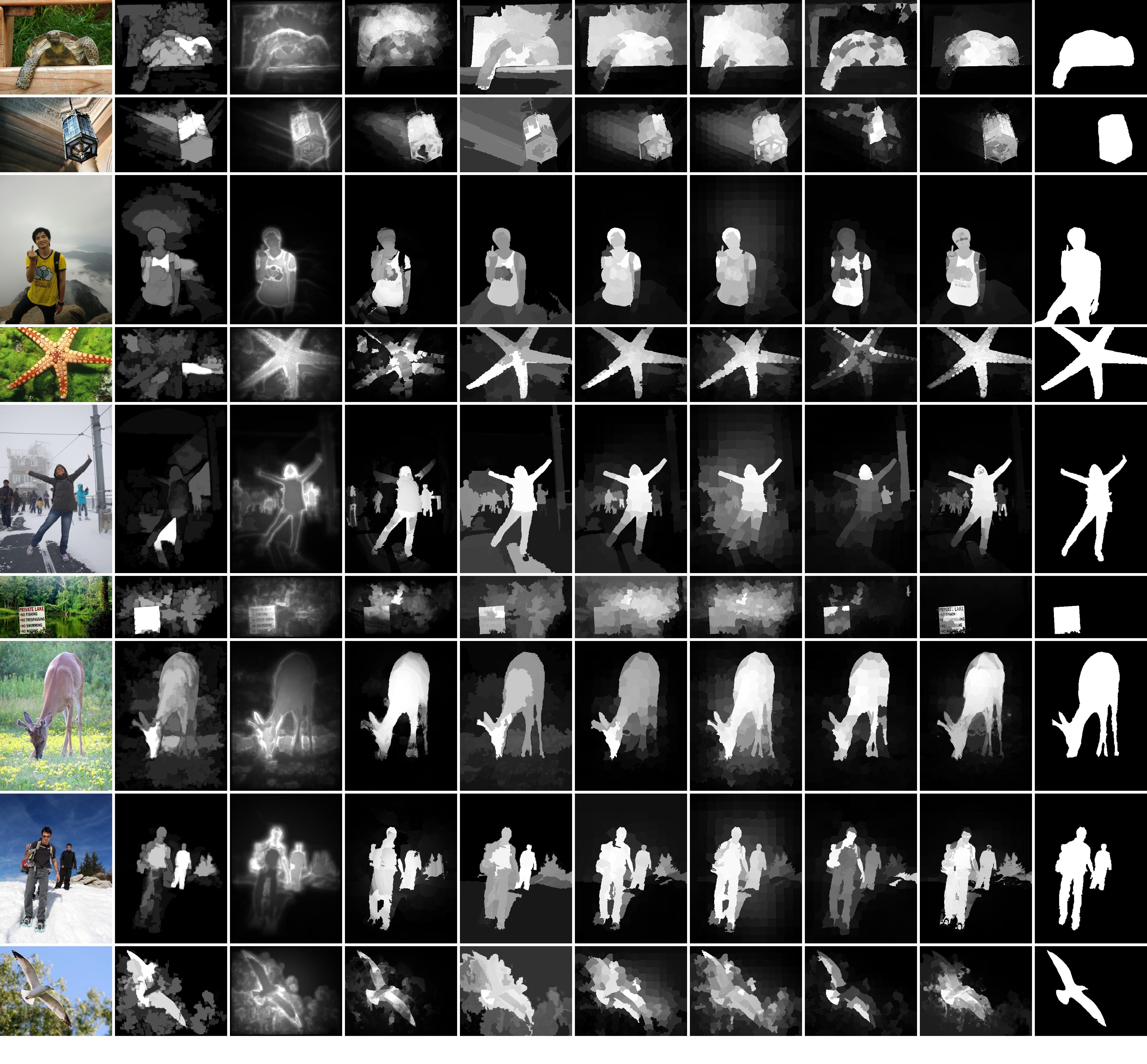}
  \put(1, .7){
  \small{
  ~Source~  ~~~~~ BMS\cite{Boolean} ~~~~~ PD\cite{PatchDistinct13}  ~~~~~~~DS\cite{PCA_dlut}  ~~~~~~~HS\cite{Hierarchy}  ~~~~~~ MR\cite{Manifold} ~~~~~~MK\cite{Markov_dlut}  ~~~~~~HPS\cite{Hyper13}  ~~~~~~ LPS     ~~~~~~~~~ GT
  }
  }
  \end{overpic}
  \vspace{-.2cm}
  \caption{ Visual comparison of previous methods, our algorithm (LPS) and ground truth (GT) on the CCSD-1000 dataset.
    The examples in the last two rows show failure cases where LPS abundantly detects more `salient area' or powerlessly segments the salient object from the complex background.
    %For more results, see the supplementary.
    %See supplementary for comparison results of other approaches.
  } \label{fig:cmp}
  \vspace{-.4cm}
\end{figure*}

\begin{figure*}
  \centering
  \begin{overpic}[width=\textwidth]{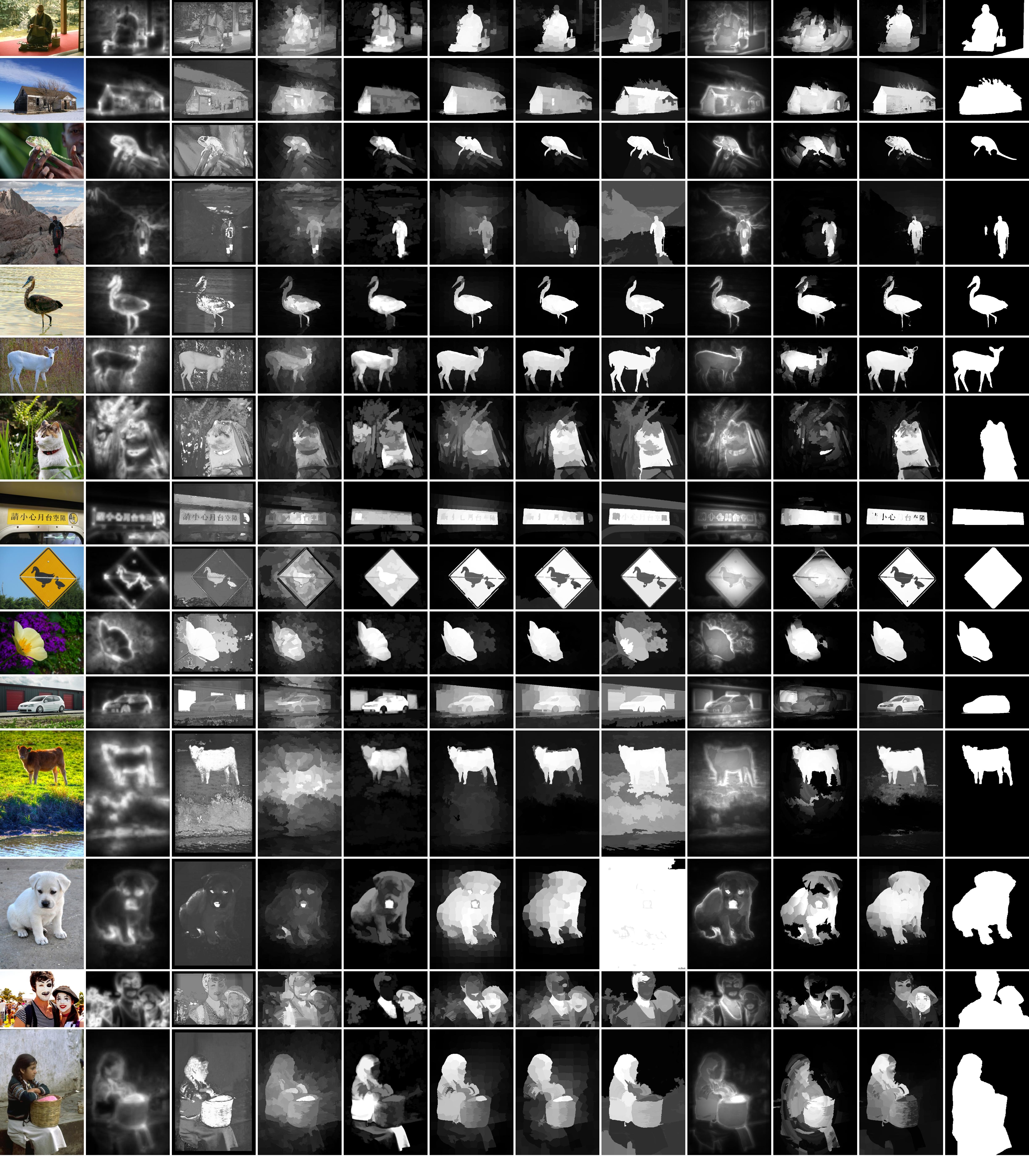}
  \put(.1, .1){ \small
  ~Source   ~~~~ ~CA\cite{10cvpr/goferman_context} ~~~~ RA\cite{Finland10}   ~~~~LR\cite{Northwestern12} ~~~~BMS\cite{Boolean}
  ~~~MK\cite{Markov_dlut}   ~~~MR\cite{Manifold} ~~~~HS\cite{Hierarchy}  ~~~~PD\cite{PatchDistinct13} ~~~~ DS\cite{PCA_dlut} ~~~~~LPS     ~~~~~~~ GT
  }
  \end{overpic}
  \caption{Visual comparison of previous methods, our algorithm (LPS) and ground truth (GT) on the PASCAL-S and THUS-10,000 dataset.
  The examples in the last two rows show failure cases where LPS carelessly misses the foreground parts that belong to the salient people.
  } \label{fig:thus}
\end{figure*}

Examples in the last rows of Fig.\ref{fig:msra_5k} to Fig.\ref{fig:thus} show failure cases where  the proposed algorithm is unable to detect
the salient object in some scenarios.
Currently we only use the color  information to construct the affinity matrix because the structure description
of an image is included in the pre-abstraction processing. As shown in Sec.\ref{sec:co-trans},
the %commonly used
structure based descriptor does not work well due to redundant extraction of the foreground and noisy
extraction of the background. However, we believe that investigating more sophisticated feature representations for the co-transduction  algorithm would be greatly beneficial.
It would also be interesting to exploit top-down and category-independent semantic information to enhance the current results.
We will leave these two directions as the starting point of our future research.

\section{Conclusions}

In this work, we explicitly propose a label propagation method in salient object detection.
For some images, an inner label propagation via boundary labels alone  obtains good
visual and evaluation results; for more natural and complex images in the wild, a co-transduction algorithm
which combines boundary superpixels with objectness labels %extracted from a 3-cue-center-biased objectness score,
 can   have better saliency assignment.
The compactness criterion decides whether the final saliency map is
simply a production of the inner propagation or a fusion outcome of the inter propagation.
The proposed method achieves  superior performance in terms of different evaluation metrics,  compared with
the state-of-the-arts on five benchmark image datasets.

\bibliographystyle{IEEEtran}
\bibliography{saliency_ECCV14}

\end{document}